\pdfoutput=1

\documentclass[11pt]{article}

\usepackage[final]{acl}

\usepackage{times}
\usepackage{latexsym}
\usepackage{graphicx}
\usepackage{enumitem}

\usepackage[T1]{fontenc}

\usepackage[utf8]{inputenc}

\usepackage{microtype}

\usepackage{inconsolata}

\usepackage{graphicx}
\usepackage{multirow}
\usepackage{booktabs}
\usepackage{longtable}
\usepackage{array}
\usepackage{float}

\title{Kahaani: A Multimodal Co-Creative Storytelling System}

\author{
Samee Arif\textsuperscript{\rm 1}, \textsuperscript{\rm *}Muhammad Saad Haroon\textsuperscript{\rm 1}, \textsuperscript{\rm *}\textbf{Aamina Jamal Khan}\textsuperscript{\rm 1},\\ \textbf{Taimoor Arif}\textsuperscript{\rm 2}, \textbf{Agha Ali Raza}\textsuperscript{\rm 1}, \textbf{Awais Athar}\textsuperscript{\rm 3} \\
\textsuperscript{\rm 1}Lahore University of Management Sciences, \\
\textsuperscript{\rm 2}University of Michigan, \\
\textsuperscript{\rm 3}Strategize Labs \\
\texttt{\{samee.arif, 25100147, 25100162, agha.ali.raza\}@lums.edu.pk}\\
\texttt{taimoora@umich.edu}, \texttt{awais@strategize.inc}
}

\begin{document}
\maketitle
\renewcommand{\thefootnote}{\fnsymbol{footnote}}
\footnotetext[1]{These authors contributed equally to this work.}
\renewcommand{\thefootnote}{\arabic{footnote}}
\begin{abstract}
This paper introduces Kahaani, a multimodal, co-creative storytelling system that leverages Generative Artificial Intelligence, designed for children to address the challenge of sustaining engagement to foster educational narrative experiences. Here we define co-creative as a collaborative creative process in which both the child and Kahaani contribute to the generation of the story. The system combines Large Language Model (LLM), Text-to-Speech (TTS), Text-to-Music (TTM), and Text-to-Video (TTV) generation to produce a rich, immersive, and accessible storytelling experience. The system grounds the co-creation process in two classical storytelling framework, Freytag’s Pyramid and Propp’s Narrative Functions. The main goals of Kahaani are: (1) to help children improve their English skills, (2) to teach important life lessons through story morals, and (3) to help them understand how stories are structured, all in a fun and engaging way. We present evaluations for each AI component used, along with a user study involving three parent–child pairs to assess the overall experience and educational value of the system.
\end{abstract}

\section{Introduction}
\begin{figure*}[h!]
    \begin{center}
    \centerline{\includegraphics[width=\textwidth]{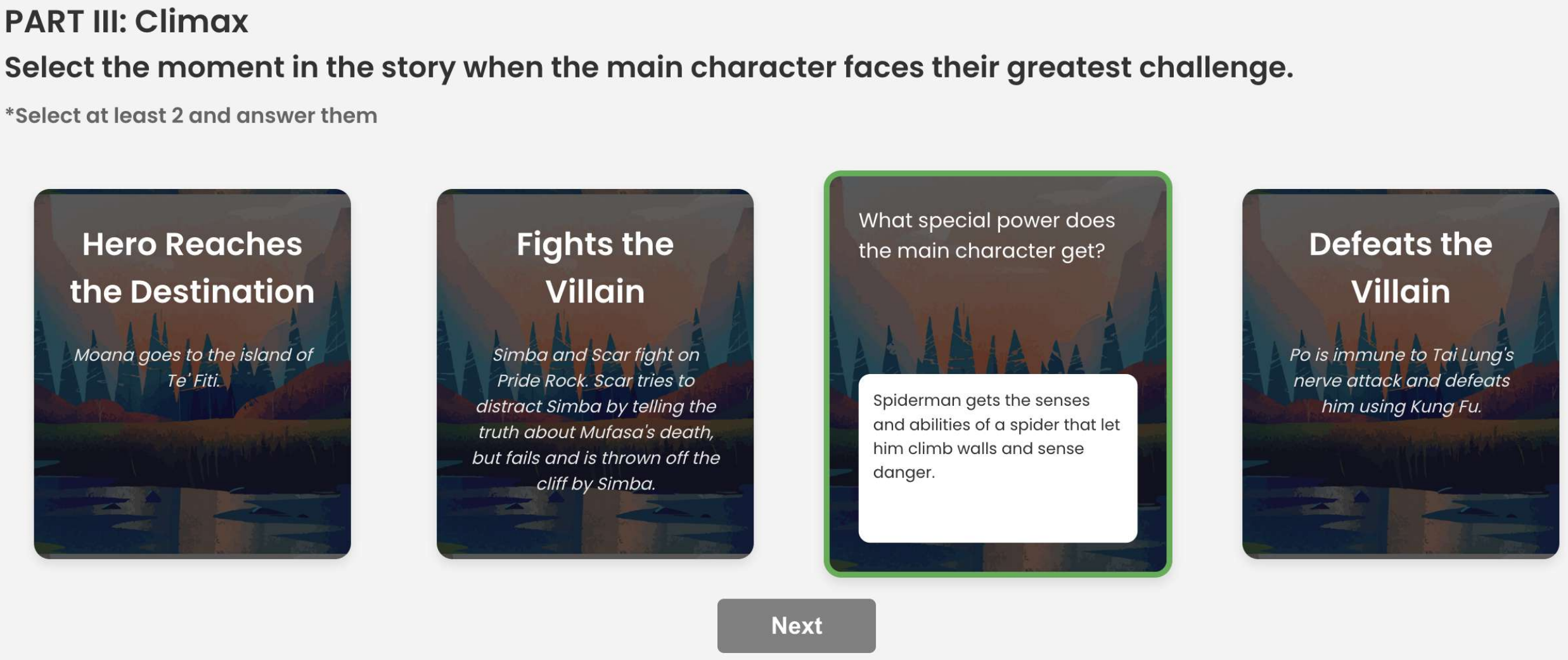}}
    \caption{\small{The system's front-end design.}}
    \label{fig:system-screen}
    \end{center}
    \vskip -0.3in
\end{figure*}

In this paper, we introduce Kahaani\footnote{\url{https://github.com/SaadH-077/kahaani}}, a system designed to co-create multimodal stories for children. The central problem Kahaani aims to address is the difficulty children often have in maintaining attention and developing narrative competence with traditional story delivery. To support children in overcoming these challenges, the system is structured around two well-established storytelling frameworks: Freytag's Pyramid from his book \textit{The technique of Drama}\footnote{\url{https://www.gutenberg.org/files/50616/50616-h/50616-h.htm}} and Propp's 31 narrative functions from his book \textit{Morphology of the Folktale}\footnote{\url{https://www.jstor.org/stable/10.7560/783911}}. Freytag's Pyramid provides a foundational structure, dividing stories into five phases, exposition, rising action, climax, falling action, and resolution, ensuring that the generated narratives follow a coherent and engaging progression. Propp's functions serve as a guide for generating the essential elements found in folktales, offering a consistent framework for creating dynamic and imaginative stories that resonate with young audiences. These frameworks can also effectively enhance storytelling skills \citep{article-support}. From a pedagogical perspective, these frameworks allow children to actively participate in the storytelling process rather than passively consuming content.


The conversion of written text into spoken words provides auditory stimuli that can improve reading comprehension, especially for students with learning disabilities. Studies have shown that TTS can support struggling readers by offering audio input as digital text is read aloud, aiding in comprehension and retention \citep{ba7c3d0a0f8f4dfa9b63c17d0fea54bf}. Visual stimuli can enhance understanding and retention of information. Research indicates that combining text and images in multimedia explanations is more effective than text alone, as it caters to diverse learning preferences and helps maintain attention \citep{the-effects-of-segmentations}. Background music adds an emotional layer to narratives, facilitating deeper engagement and supporting memory formation. Music has been recognized for its ability to evoke emotions and create a captivating environment, which can enhance memory retention by increasing engagement \citep{essay95283}. By integrating these components, our system addresses the limitations of traditional storytelling methods that often lack multimodal stimuli, leading to reduced attention spans and engagement. This multimodal approach ensures that storytelling is not only entertaining but also educational, catering to various learning styles and promoting cognitive development.

\section{System Architecture}
\begin{figure}[h!]
    \begin{center}
    \centerline{\includegraphics[width=\columnwidth]{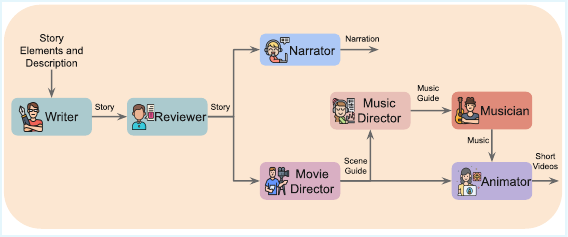}}
    \caption{\small{Multi-agent system for multimodal story generation.}}
    \label{fig:system-architecture}
    \end{center}
    \vskip -0.3in
\end{figure}

Figure \ref{fig:system-architecture} shows the architecture of our application. The storytelling process is divided into five phases, reflecting Freytag's pyramid, with Propp's narrative functions (e.g., Interdiction, Villainy and Absentation) distributed throughout these phases. In each phase, the user selects cards that represent the relevant Propp functions and answers specific questions based on the chosen function. The system's input design is shown in Figure \ref{fig:system-screen}.

Once the five phases are completed, the input is passed to the Writer. The Writer LLM generates a story based on the user’s inputs. The generated story is then reviewed by the Reviewer LLM. The Reviewer checks whether the story is suitable for children, making any necessary edits to ensure it is age-appropriate. After the story is finalized, it is sent to the Narrator (TTS), which converts the text into natural speech. Simultaneously, the story is also forwarded to the Film Director LLM, which writes a detailed scene guide for each paragraph, outlining the visual elements needed to represent the narrative. For each paragraph’s scene guide, the Music Director LLM generates a music guide that reflects the emotional tone and context of the story. This guide is passed to the Musician (TTM), which creates the corresponding background music. Both the scene guide and the music are then used by the Animator (TTV) to generate a video for a cohesive and immersive animation. The final outputs include the textual story, the narrated audio, and the animated video with background music, creating a multimodal storytelling experience for children.

\section{Related Work}
The survey paper by \citet[]{alhussain} provides a systematic review of various methods in automatic story generation, detailing the evolution from rule-based to AI-driven approaches, and discusses the challenges and future directions in enhancing narrative creativity and coherence. \citet[]{DBLP:journals/corr/abs-1805-04833} introduced a method that first generates an outline for a story, followed by the story itself. They used a dataset of user-submitted prompts and user-generated stories for this method. Our methods build on this work by replacing the user who generates the stories with an LLM to improve the speed and scale of the data generated. PlotMachines \citet{DBLP:journals/corr/abs-2004-14967} is a similar system that takes the outline of the story as a set of phrases and generates the story text based on those. The work by \citet[]{xie2024creatingsuspensefulstoriesiterative} takes an iterative approach to story planning. They ask the system a bunch of questions that set up the outline/premise of the story and then generate the story once these details are ironed out. SWAG \citet{patel2024swagstorytellingactionguidance} is another story-generation framework that uses two models, one that generates the story, and the second that guides the story-generation process. Multiple papers (\citet[]{zhu2023endtoendstoryplotgenerator}, \citet[]{huang2023conveyingpredictedfutureusers}, \citet[]{ma-etal-2023-coherent}, \citet[]{jin2022plotwritingpretrainedlanguage}) are specifically targeted toward plot generation and development, which is the focal point of delivering a good story, on top of the structural and linguistic requirements. These methods include step-by-step generation of important parts of the story \citet{zhu2023endtoendstoryplotgenerator}, generating and utilizing a sequence of events from a fine-tuned model \citet{ma-etal-2023-coherent}, and specific tasks using pre-trained models \citet{jin2022plotwritingpretrainedlanguage}. 

While existing story-generation systems focus primarily on text-based plot development and iterative planning, they largely omit multimodal, co-creative frameworks that integrate narrative text with audio, animation, and music for educational and engagement-focused applications, which is the gap Kahaani aims to fill.


\section{Experimental Setup}
We conduct human evaluation of LLM for story generation, TTS narration, TTV animation. The human evaluators for each module were six Computer Science students with English as the medium of education. Each criterion was assigned a score of of 0, 1, or 2 (breakdown given in Appendix Section \ref{sec:score-breakdown}). After evaluating the components individually, we selected the top-performing models for each of the other modules and used them to conduct the user study of the complete system.

\subsection{LLMs for Story Generation}
To evaluate the LLMs listed in Table \ref{tab:models} for story generation, we collected a test dataset. We created a Survey form divided into five parts based on Freytag's Pyramid. Propp's 31 narrative functions were distributed across these sections to align with each part of the narrative structure. Undergraduate students, with English as the medium of education, were asked to fill out the form. In total, we collected 50 test prompts for evaluation.

\begin{table}[h]
\centering
\small
\begin{tabular}{l@{\hspace{0.3cm}}l}
\toprule
\textbf{Category} & \textbf{Models} \\ \midrule
\multirow{2}{*}{Small-Scale LLM}
  & Llama-3.1-8b \\
  & Gemma-2-9b \\[.5em]
\multirow{2}{*}{Mid-Scale LLM}
  & Gemma-2-27b \\
  & Llama-3.1-70b \\[.5em]
\multirow{2}{*}{Large-Scale LLM}
  & GPT-4o-mini (2024-07-18) \\
  & GPT-4o (2024-05-13) \\
\bottomrule
\end{tabular}
\caption{Categories of LLMs used in the study.}
\label{tab:models}
\vskip 0in
\end{table}

In order to understand how teachers evaluate stories submitted by their students, we conducted semi-structured interviews with 3 teachers who teach English as a subject to students of age group 6 to 12. The teachers emphasized assessing the content of the story for its quality. This includes evaluating the language used, checking the logical sequence of the story, usage of “key terms” to “cue the reader to the direction of the discourse”. The teachers noted:
\begin{center}
    \textit{“.....we try not to use any harmful/abusive language”\\
    “We check if there is a clear beginning and end”\\
    “Does the writer stick to the topic of the story?”}
\end{center}
Other than the content, the structure of the story is also scrutinized. One of the teachers said:
\begin{center}
    \textit{“A group of words does not make a complete sentence. Sentences that are made up of a group of words without a direction need to be penalized”}
\end{center}

Keeping prior literature and our interviews in mind, we use the following framework to evaluate our stories.
\begin{enumerate}[noitemsep, topsep=0pt]
    \item \textbf{Grammar:} Is the story grammatically correct? \citep{Guan_Wang_Huang_2019}
    \item \textbf{Linguistic Consistency:} Is the story consistent in its language? Language consistency means using words of a similar nature and difficulty across the whole story. \citep{Roemmele2017EvaluatingSG}
    \item \textbf{Appropriate Language:} Does the story use appropriate language, and are offensive or inappropriate words avoided? \citep{bhandari-brennan-2023-trustworthiness}
    \item \textbf{Structural Consistency:}  Is the story consistent in its structure? Structural consistency means that sentences follow a similar structure and are connected with little to no disjoint. \citep{10.1609/aaai.v33i01.33017378}
    \item \textbf{Creativity:} Is the story interesting and enjoyable? Does it capture your attention and make you want to keep reading? \citep{pascual-etal-2021-plug-play}
    \item \textbf{Adherence To Instructions:} Does the story adhere to the prompt parameters passed by the user? \citep{peng-etal-2018-towards}
    \item \textbf{Naturalness:} Does the generated story seem like it is written by a human? \citep{pascual-etal-2021-plug-play}
\end{enumerate}

\subsection{LLMs for Content Moderation}
The goal of this evaluation is to ensure that children are not exposed to inappropriate content within stories. To assess the effectiveness of the LLMs as content reviewers, We gathered a diverse dataset of 100 stories from Project Gutenberg, 50 appropriate and 50 inappropriate as rated by our evaluators. Additionally, we generated a dataset of 50 stories using LLMs, which were similarly labeled as appropriate or inappropriate. These stories were labeled by human annotators based on the presence of violent, explicit, or otherwise unsuitable material for children. This labeled dataset was used to evaluate the LLMs in Table \ref{tab:models}. The system prompt for this task is given in Appendix Section \ref{sec:prompts}.

\subsection{Text-to-Speech Models}
We evaluate XTTS-v2\footnote{\url{https://huggingface.co/coqui/XTTS-v2}} and StyleTTS 2 \citep{li2023styletts2humanleveltexttospeech}, top two open source models on TTS leader board on Hugging Face\footnote{\url{https://huggingface.co/spaces/TTS-AGI/TTS-Arena}}. We sample 50 random paragraphs from free stories available on Project Gutenberg\footnote{\url{https://www.gutenberg.org/}}. Additionally, we use two reference audios (one narrated by a female speaker and one by a male speaker) to perform voice cloning. Both speakers are computer science researchers with high proficiency in English. We run inference for both speakers across both TTS models and compare their performance based on the following criteria:
\begin{enumerate}[noitemsep, topsep=0pt]
    \item \textbf{Clarity and Pauses:} The speech should be easily understandable with minimal listening effort and should have well-placed pauses.
    \item \textbf{Information and Emotion Preservation:} The context, and emotions conveyed in the text should be accurately reflected in the audio, for the immersive storytelling experience.
    \item \textbf{Intonation and Naturalness:} The changes in pitch, amplitude, and stress should feel natural, resulting in smooth, realistic speech flow.
    \item \textbf{Fluency and Pronunciation:} Pronunciation should be accurate, fluent, and clear, ensuring proper word distinction and avoiding misinterpretation.
\end{enumerate}
These criteria were inspired by the evaluation protocol proposed by \citet{Hinterleitner2011AnEP} and \citet{9688073}. Additionally, for Voice Cloning, the raters provided a score of 0, 1, or 2 for the overall model performance, assessing how closely the synthesized voice matched the reference voice.

\subsection{Text-to-Video Models}
For the TTV module we evaluate CogVideoX-5b \citep{yang2024cogvideox} on three types of animation styles: cartoon, anime and animated. The `animated` style gives the model freehand to generate visuals in any animation style. We sample 50 random paragraphs from the Project Gutenberg free stories and pass them though the Director agent to get the scene guide. The system prompt for Director is given in Appendix \ref{sec:prompts}. The scene guide is then used for video generation. For each video, we assess the following metrics:
\begin{enumerate}[noitemsep, topsep=0pt]
    \item \textbf{Naturalness Assessment:} There should not be any anomalies or odd behaviors, such as unnatural movements or deformations \citep{liao2024evaluationtexttovideogenerationmodels}.
    \item \textbf{Temporal Quality:} This evaluates how smoothly and coherently the video transitions from one frame to another \citep{liu2023fetvbenchmarkfinegrainedevaluation}.
    \item \textbf{Fine-Grained Alignment:} This metric focuses on the alignment of specific video attributes like color, speed, or motion direction \citep{liu2023fetvbenchmarkfinegrainedevaluation}.
    \item \textbf{Overall Alignment:} This measures how well the video matches the content described in the scene guide \citep{liu2023fetvbenchmarkfinegrainedevaluation}, \citep{wu2024bettermetrictexttovideogeneration}.
    \item \textbf{Child-Friendliness:} This metric assesses whether the video content is appropriate for children.
\end{enumerate}

\subsection{User Study}
To evaluate the overall experience and educational value of the system, we conducted a user study with three parent-child pairs. Each child interacted with the system to co-create a story, and both children and their parents were asked to provide feedback through structured questionnaires. The goal was to assess the system’s usability, engagement, age-appropriateness, and educational impact from both perspectives. To make it easier for children to respond, we included mostly rating-based questions (represented with a * below) on a 1–5 scale. \\

\textbf{Children’s Questionnaire}
\begin{enumerate}[noitemsep, topsep=0pt]
    \itemsep0em
    \item How easy was the story to understand?*
    \item How likely are you to recommend this system to your friends?*
    \item How much did you like the animations?*
    \item How much did you like the way the story was told (e.g., narration, voice)?*
    \item How much would you rate the generated story itself?*
    \item Overall, how would you rate your experience with the storytelling system?*
    \item Did you learn something new from the story? What was it?
    \item What changes would you suggest to improve the overall experience?
\end{enumerate}
\vspace{0.5em}
\textbf{Parents’ Questionnaire}
\begin{enumerate}[noitemsep, topsep=0pt]
    \itemsep0em
    \item How appropriate was the content for your child’s age?*
    \item How engaged was your child while using the system?*
    \item How likely are you to recommend this system to other parents?*
    \item How much did you like the system’s design (e.g., visuals, ease of use)?*
    \item Overall, how satisfied are you with the storytelling system?*
    \item How did your child react to using the storytelling system?
    \item Do you think the system helped your child learn something new?
    \item Do you think this system will have any improvement in your child’s language skills or creativity?
\end{enumerate}

\section{Results \& Discussion}
\subsection{LLMs for Story Generation}
Table \ref{tab:win-lose-tie-rates-llm} summarizes the pairwise comparisons among Gemma-2-9b, Gemma-2-27b \citep{google2024gemma2}, Llama-3.1-8b, Llama-3.1-70b \citep{meta2024llama3}, GPT-4o, and GPT-4o-mini \citep{openai2024gpt4} based on human evaluations of their story outputs.

\begin{center}
\begin{table}[h!]
\centering
\renewcommand{\arraystretch}{1}
\small
\begin{tabular}{llcccccc}
\toprule
& & \multicolumn{3}{c}{\textbf{Rates (\%)}} \\
\cmidrule(lr){3-5}
\textbf{Test Model} & \textbf{Versus Model} & \textbf{Win} & \textbf{Tie} & \textbf{Loss} \\
\midrule
Gemma-9b & Gemma-27b & 39.67 & 34.67 & 25.67 \\
Gemma-9b & Llama-8b & 39.17 & 34.00 & 26.83 \\
Gemma-9b & Llama-70b & 37.44 & 35.33 & 27.22 \\
Gemma-9b & GPT-4o & 40.00 & 33.42 & 26.58 \\
Gemma-9b & GPT-4o-mini & 38.27 & 33.60 & 28.13 \\
Gemma-27b & Llama-8b & 31.33 & 32.33 & 36.33 \\
Gemma-27b & Llama-70b & 28.33 & 36.00 & 35.67 \\
Gemma-27b & GPT-4o & 31.56 & 33.67 & 34.78 \\
Gemma-27b & GPT-4o-mini & 29.92 & 33.08 & 37.00 \\
Llama-8b & Llama-70b & 28.33 & 41.00 & 30.67 \\
Llama-8b & GPT-4o & 34.00 & 38.17 & 27.83 \\
Llama-8b & GPT-4o-mini & 30.33 & 37.78 & 31.89 \\
Llama-70b & GPT-4o & 43.33 & 29.00 & 27.67 \\
Llama-70b & GPT-4o-mini & 34.00 & 33.67 & 32.33 \\
GPT-4o & GPT-4o-mini & 16.33 & 36.67 & 47.00 \\
\midrule
\end{tabular}
\caption{\small{Win-rate, tie-rate, and loss-rate for Test Model against Versus Model based on scoring by all raters.}}
\label{tab:win-lose-tie-rates-llm}
\end{table}
\vskip -0.35in
\end{center}

\begin{figure*}[h]
    \begin{center}
    \centerline{\includegraphics[width=\textwidth]{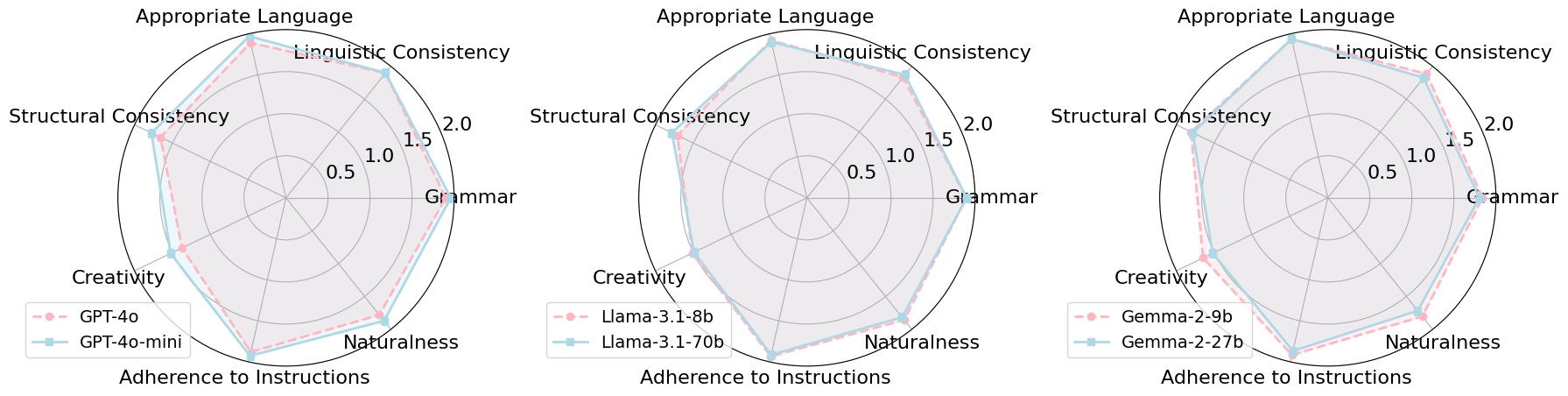}}
    \caption{\small{Comparison of the six LLMs for each metric based on average human-evaluation scores.}}
    \label{fig:llm-comparison}
    \end{center}
    \vskip -0.3in
\end{figure*}

A notable observation is that simply scaling up parameter counts does not guarantee stronger performance in the downstream task of story generation. Gemma-2-9b outperforms its larger counterpart Gemma-2-27b (39.67\% vs.\ 25.67\% in win-rate), underscoring that bigger parameter counts are not always decisive for story-generation quality. Likewise  Llama-3.1-70b achieves notable 43.33\% wins and only 27.67\% losses against GPT-4o. GPT-4o also performs poorly against GPT-4o-mini with only 16.33\% win-rate. Across most pairwise comparisons, tie-rates hover near 30\%, suggesting that many of these models are closely matched in generating coherent and engaging stories. Based on these head-to-head outcomes, Gemma-2-9b emerges as the most consistently strong performer overall, outperforming all the other models. Interestingly, the bigger model from the same family, Gemma-2-27b is the weakest based on the human evaluation, wining none of the pairwise comparisons.

In Table \ref{tab:bt_ranking}, we use the Bradley–Terry model \citep{JSSv012i01} to rank the LLMs based on their pairwise comparison results, with ties counted as half-wins. The top-performing model for our generative task is Gemma-9b, followed by GPT-4o-mini. Interestingly, for both Gemma and GPT, the smaller models outperform their larger counterparts, whereas for Llama, the larger Llama-70b slightly outperforms the smaller Llama-8b. Overall, these results suggest that smaller models tend to excel at creative and linguistic tasks. While larger models offer longer context windows and improved accuracy, this has a tradeoff against long-form, creative generation tasks.

\begin{table}[H]
\centering
\renewcommand{\arraystretch}{1.2}
\small
\begin{tabular}{ccc}
\toprule
\textbf{Rank} & \textbf{Model} & \textbf{BT Strength ($\pi_i$)} \\
\midrule
1 & Gemma-9b    & 1.224 \\
2 & GPT-4o-mini & 1.097 \\
3 & Llama-70b   & 1.058 \\
4 & Llama-8b    & 0.983 \\
5 & Gemma-27b   & 0.884 \\
6 & GPT-4o      & 0.810 \\
\bottomrule
\end{tabular}
\caption{Bradley-Terry \citep{JSSv012i01} ranking of models based on pairwise comparisons, with ties counted as half-wins. Higher $\pi_i$ indicates a stronger model.}
\label{tab:bt_ranking}
\end{table}

Figure \ref{fig:llm-comparison} presents average human-evaluation scores (on a 0–2 scale) across seven criteria—Grammar, Linguistic Consistency, Appropriate Language, Structural Consistency, Creativity, Adherence to Instructions, and Naturalness. A closer look at Grammar scores shows that GPT-4o-mini leads with an average of 1.96, while most other models hover around 1.80–1.90. For Linguistic Consistency, GPT-4o and GPT-4o-mini both top out at 1.90, whereas Llama-3.1-8b, Llama-3.1-70b, and Gemma-2-9b trail slightly in the 1.83–1.88 range, suggesting all models are fairly consistent in using language of similar register or difficulty throughout a story. In Appropriate Language, GPT-4o-mini again scores highest (1.97), closely followed by the Gemma-2 series at 1.94, signaling strong ability to avoid potentially offensive or overly complex wording.

In Structural Consistency, Gemma-2-9b (1.81) stands out as the most coherent, edging out Gemma-2-27b, GPT-4o-mini, and Llama-3.1-70b (all near 1.78) while GPT-4o (1.66) lags notably. In Creativity, Gemma-2-9b achieves the top mean score (1.65), surpassing all other models—especially GPT-4o at 1.37—indicating Gemma-2-9b’s stronger capacity for generating engaging stories. Under Adherence to Instructions, Llama-3.1-8b and GPT-4o-mini both earn the highest score (1.93), followed closely by Gemma-2-9b (1.92) and Llama-3.1-70b (1.91), showcasing that all models tend to follow user prompts precisely. Lastly, Naturalness is also led by GPT-4o-mini at 1.87, with Llama-3.1-8b just behind (1.86), reflecting how “human-like” or fluid their story outputs appear to human readers.

Overall, GPT-4o-mini emerges as most consistent across the board, consistently appearing at or near the top in most categories, especially Grammar, Appropriate Language, and Naturalness. Gemma-2-9b scores highest in Creativity and Structural Consistency, making it appealing for users. In contrast, GPT-4o sits at the lower end in several metrics particularly Creativity and Structural Consistency.

\subsection{LLMs for Content Moderation}
Table \ref{tab:llm-moderation} shows the False Positive Rate (FPR) for content classification for each model. Our aim is to minimize the FPR because our priority is to avoid mistakenly labeling inappropriate stories as appropriate, as this could result in children being exposed to unsuitable content. Ensuring a low FPR is essential for maintaining a safe and child-friendly storytelling environment. GPT-4o achieves the lowest FPR, with 9\% FPR for the Project Gutenberg dataset and 0\% FPR for the LLM-generated story dataset.

The Project Gutenberg dataset contains human-written stories, where the risk of inappropriate content is higher. However, when generating content, LLMs have built-in safeguards \citep{wang-etal-2024-answer} that significantly reduce the probability of producing inappropriate material, making the likelihood of harmful content in generated stories very low.

\begin{table}[H]
\centering
\small
\begin{tabular}{lcc}
\toprule
\textbf{Models} & \textbf{Gutenberg} & \textbf{Synthetic} \\ \midrule
GPT-4o-mini & 26 & 25 \\
GPT-4o & 9 & 0 \\
Llama-3.1-8b & 39 & 37.5 \\
Llama-3.1-70b & 32 & 37.5 \\
Gemma-2-9b & 34 & 0 \\
Gemma-2-27b & 42 & 12.5 \\
\bottomrule
\end{tabular}
\caption{\small{False Positive Rate (FPR) for content classification.}}
\label{tab:llm-moderation}
\vskip -0.1in
\end{table}

\subsection{Text-to-Speech}
Table \ref{tab:win-lose-tie-rates-tts} shows how the two text-to-speech models, XTTSv2 and StyleTTS 2, fared under female and male reference voices. When using the female reference audio, XTTSv2 achieves a 37.83\% win-rate against StyleTTS 2 but also has a slightly higher loss-rate of 39.00\%, reflecting a very close quality. By contrast, with the male reference audio, XTTSv2 performs somewhat better overall (39.67\% win vs. 34.67\% loss). Tie-rates remain in the 23–26\% range for both scenarios, indicating that both models produce similarly acceptable outputs in a substantial fraction of cases. Although these comparisons do not reveal a single dominant TTS system across all conditions, the trends suggest that XTTSv2 may slightly outperform StyleTTS 2 in handling the male voice, whereas StyleTTS 2 holds a marginal edge when cloning the female voice. 

\begin{center}
\begin{table}[ht!]
\centering
\renewcommand{\arraystretch}{1}
\small
\begin{tabular}{llccccc}
\toprule
& & \multicolumn{3}{c}{\textbf{Rates (\%)}} \\
\cmidrule(lr){3-5}
\textbf{Test Model} & \textbf{Versus Model} & \textbf{Win} & \textbf{Tie} & \textbf{Loss} \\
\midrule
XTTSv2 & StyleTTS 2 & 37.83 & 23.17 & 39.00 \\
XTTSv2 & StyleTTS 2 & 39.67 & 25.67 & 34.67 \\
\bottomrule
\end{tabular}
\caption{\small{Win-rate, tie-rate, and loss-rate for TTS 1 against TTS 2 based on scoring by both raters. Row 1 is for female reference audio and row 2 is for male reference audio.}}
\label{tab:win-lose-tie-rates-tts}
\end{table}
\vskip -0.3in
\end{center}

\begin{figure}[t!]
    \begin{center}
    \centerline{\includegraphics[width=\columnwidth]{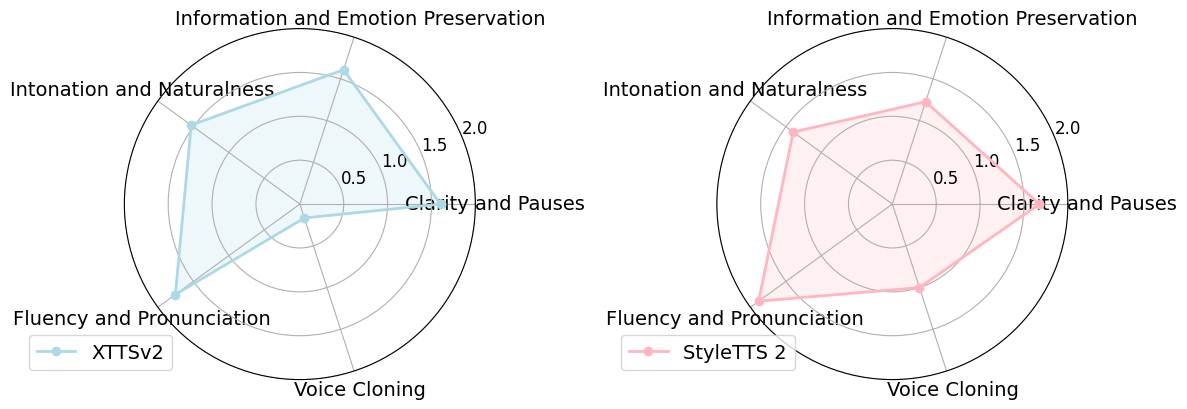}}
    \caption{\small{Comparison of XTTSv2 and StyleTTS 2 (female voice) for each metric.}}
    \label{fig:female-tts-comparison}
    \end{center}
    \vskip -0.4in
\end{figure}

Figure \ref{fig:female-tts-comparison} shows a category-wise breakdown of each model’s average TTS scores using a female reference voice. When evaluated with a female reference voice, StyleTTS 2 achieves higher scores in Clarity and Pauses (1.68 vs. 1.60) as well as Fluency and Pronunciation (1.88 vs. 1.75), suggesting it provides more precise articulation and smoother pacing overall. Although StyleTTS 2 performs better in Voice Cloning (1.00 vs. 0.17), it score is still low. By contrast, XTTSv2 scores notably higher in Information and Emotion Preservation (1.61 vs. 1.23), implying it better conveys the emotional cues. In terms of Intonation and Naturalness, the two models are relatively close (1.53 vs. 1.40), though XTTSv2’s slightly higher score points to marginally more human-like pitch variation. Overall, these results suggest that StyleTTS 2 may be the stronger choice when precise pronunciation and voice replication are prioritized, whereas XTTSv2 appears better suited for scenarios that demand heightened emotional expressiveness.

\begin{figure}[h!]
    \begin{center}
    \centerline{\includegraphics[width=\columnwidth]{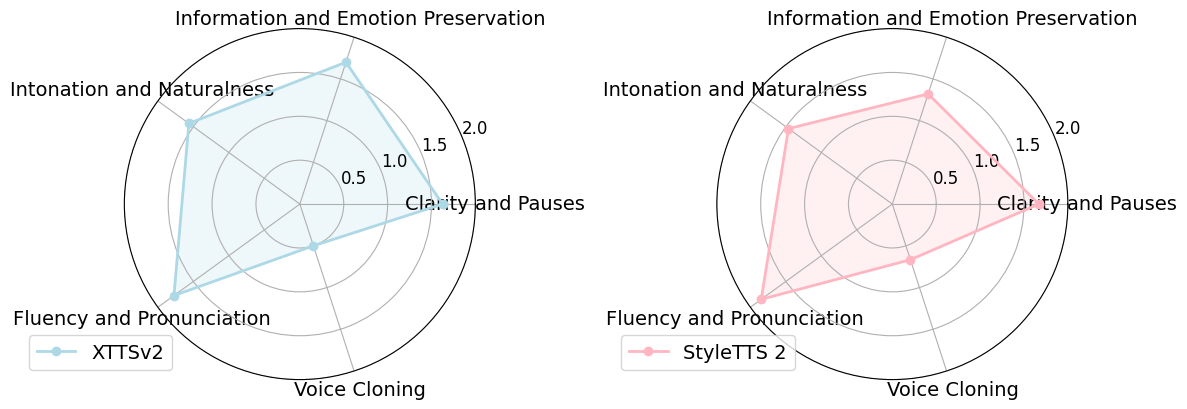}}
    \caption{\small{Comparison of XTTSv2 and StyleTTS 2 (male voice) for each metric.}}
    \label{fig:male-tts-comparison}
    \end{center}
    \vskip -0.3in
\end{figure}

Figure \ref{fig:male-tts-comparison} provides a category-wise breakdown of each model’s average scores for text-to-speech tasks using a male reference voice. XTTSv2 and StyleTTS 2 both demonstrate relatively high average scores in Clarity and Pauses (1.63 vs. 1.67), reflecting minimal listening effort and well-paced output. However, XTTSv2 scores noticeably higher in Information and Emotion Preservation (1.70 vs. 1.32) and Intonation and Naturalness (1.56 vs. 1.46), suggesting it captures more nuanced expressive cues and conveys pitch and rhythm changes in a more human-like manner. In contrast, StyleTTS 2 excels in Fluency and Pronunciation (1.85 vs. 1.77) and Voice Cloning (0.67 vs. 0.50). Overall, the choice between these two TTS systems lies on whether rich emotional conveyance (XTTSv2) or speaker-identity matching (StyleTTS 2) takes priority for a particular application.

\begin{figure*}[h]
    \begin{center}
    \centerline{\includegraphics[width=\textwidth]{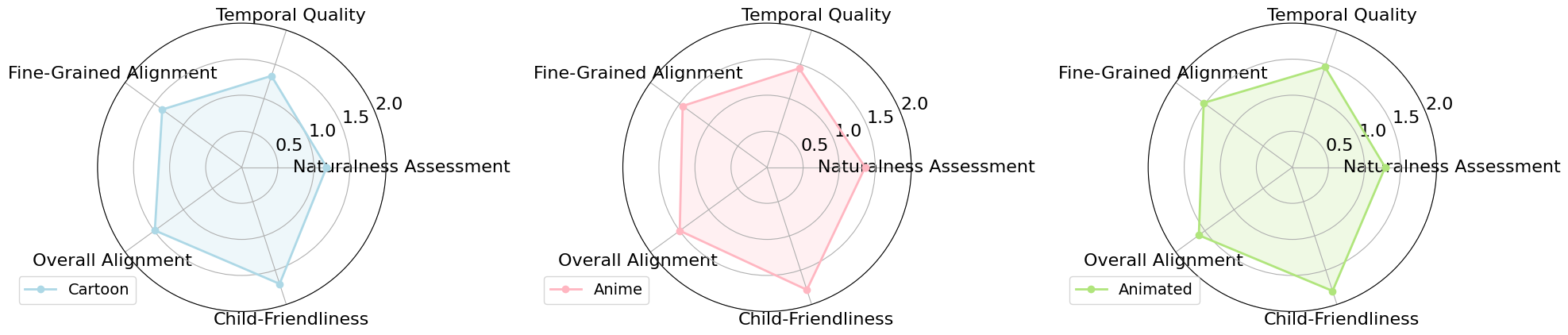}}
    \caption{\small{Comparison of three animation styles for each evaluation metric.}}
    \label{fig:ttv-comparison}
    \end{center}
    \vskip -0.3in
\end{figure*}

\subsection{Text-to-Video}
Table \ref{tab:win-lose-tie-rates-ttv} summarizes the relative performance of three text-to-video animation styles—Cartoon, Anime, and Animated—based on head-to-head comparisons. Cartoon lags behind Anime (28.67\% win vs.\ 48.33\% loss) and also against Animated (28.33\% win vs.\ 49.50\% loss), suggesting that its visuals are generally of lower quality. On the other hand, Animated leads when paired with Anime (34.33\% vs.\ 36.00\% loss), although its margin is not overwhelmingly large. Tie-rates hover around 20–30\%, reflecting that all three styles can produce some visually appealing results under certain prompts. 

\begin{center}
\begin{table}[ht!]
\centering
\renewcommand{\arraystretch}{1}
\small
\begin{tabular}{llccccc}
\toprule
& & \multicolumn{3}{c}{\textbf{Rates (\%)}} \\
\cmidrule(lr){3-5}
\textbf{Test Model} & \textbf{Versus Model} & \textbf{Win} & \textbf{Tie} & \textbf{Loss} \\
\midrule
Cartoon & Anime & 28.67 & 23.00 & 48.33 \\
Cartoon & Animated & 28.33 & 22.17 & 49.50 \\
Anime & Animated & 34.33 & 29.67 & 36.00 \\
\bottomrule
\end{tabular}
\caption{\small{Win-rate, tie-rate, and loss-rate for style 1 against style 2 based on scoring by both raters.}}
\label{tab:win-lose-tie-rates-ttv}
\end{table}
\vskip -0.4in
\end{center}

We also ranked these three models in Table \ref{tab:bt_ranking_styles} using the Bradley-Terry model \citep{JSSv012i01}, with Animated winning as the top ranked model, closely followed by Anime. Overall, Animated style appears slightly stronger in crafting engaging visuals, whereas Cartoon tends to be the weakest performer. 

\begin{table}[H]
\centering
\renewcommand{\arraystretch}{1.2}
\small
\begin{tabular}{ccc}
\toprule
\textbf{Rank} & \textbf{Style} & \textbf{BT Strength ($\pi_i$)} \\
\midrule
1 & Animated & 1.167 \\
2 & Anime    & 1.129 \\
3 & Cartoon  & 0.759 \\
\bottomrule
\end{tabular}
\caption{Bradley-Terry \citep{JSSv012i01} ranking of animation styles based on pairwise comparisons, with ties counted as half-wins. Higher $\pi_i$ indicates a stronger or more preferred style.}
\label{tab:bt_ranking_styles}
\end{table}

Figure \ref{fig:ttv-comparison} shows a category-wise breakdown of each animation style’s average scores for text-to-video tasks. When evaluating different animation styles based on five criteria, the Animated style yields the strongest Overall Alignment score (1.60) and highest Child-Friendliness (1.80). By contrast, Anime achieves the top marks for both Naturalness Assessment (1.36) and Temporal Quality (1.44), suggesting that it often offers smoother transitions and fewer visual artifacts. Although Cartoon generally scores lower in Naturalness Assessment (1.18) and Temporal Quality (1.33), it remains reasonably competitive in Overall Alignment (1.49) and offers decent Child-Friendliness (1.70). Overall, Animated stands out for aligning the generated visuals with text prompts, while Anime excels in natural-feeling motion and coherent scene transitions; Cartoon, meanwhile, appears to lag somewhat behind the other two styles, though it retains a solid score for child-centric content.

\subsection{User Study}
Based on the evaluation of each component of our system, we selected the following models: for the Writer, we use Gemma-2-9b; for the Reviewer, GPT-4o; for the Narrator, XTTSv2; GPT-4o serves as both the Movie Director and Music Director; MusicGen-Large \citep{copet2023simple} as the Musician; and CogVideoX-5b, with Animated style, is used for animation generation. This configuration ensures an optimal balance of quality, performance, and safety across all stages of the storytelling process.

\subsubsection{Children’s Responses.}
Quantitative feedback from the three children is summarized below (average scores on a 1–5 scale). The quantitative results indicate that the children found the stories generally easy to understand and enjoyable, with consistently high ratings across comprehension, overall experience, and willingness to recommend the system (all averaging 4.0). The slightly lower scores for animations (3.3) and narration quality (3.7) suggest that while the multimodal elements were engaging, there is room for improvement in the visual polish and delivery of the audio components.

\begin{enumerate}[noitemsep, topsep=0pt]
    \itemsep0em
    \item How easy was the story to understand? = \textbf{4.0}
    \item How likely are you to recommend this system to your friends? = \textbf{4.0}
    \item How much did you like the animations? = \textbf{3.3}
    \item How much did you like the way the story was told (e.g., narration, voice)? = \textbf{3.7}
    \item How much would you rate the generated story itself? = \textbf{4.0}
    \item Overall, how would you rate your experience with the storytelling system? = \textbf{4.0}
\end{enumerate}

Qualitative feedback showed that each child learned something different: one noted it was easy to \textit{“make this story by parts,”} another reflected that \textit{“you shouldn’t do things behind your parents’ back,”} and a third emphasized \textit{“learned new English words.”} For improvements, children suggested \textit{“making the final output appear more like a book-style reading story,”} \textit{“adding humor to prevent boredom,”} and \textit{“improving both animation quality and narration expressiveness.”} Overall, their responses highlight the system’s potential for engaging, age-appropriate learning while pointing to areas for refinement in visuals and narration

\subsubsection{Parents’ Responses.}
Quantitative feedback from the three parents is summarized below (average scores on a 1–5 scale). The results show that parents considered the content highly appropriate for their children (5.0) and were generally satisfied with both their child’s engagement (4.3) and the overall system experience (4.3). Willingness to recommend the system to other parents (4.0) and satisfaction with the design (3.7) received slightly lower scores, suggesting that while parents valued the educational content and engagement, there remains room for improvement in interface design and usability.

\begin{enumerate}[noitemsep, topsep=0pt]
    \itemsep0em
    \item How appropriate was the content for your child’s age? = \textbf{5.0}
    \item How engaged was your child while using the system? = \textbf{4.3}
    \item How likely are you to recommend this system to other parents? = \textbf{4.0}
    \item How much did you like the system’s design (e.g., visuals, ease of use)? = \textbf{3.7}
    \item Overall, how satisfied are you with the storytelling system? = \textbf{4.3}
\end{enumerate}

Qualitative feedback showed that children generally reacted positively to the system, though enthusiasm varied. One parent noted their child was \textit{“excited,”} another that she \textit{“enjoyed it but not as much as I thought she would, especially given she loves to write stories,”} while a third observed he \textit{“was engaged and was looking forward to what will be generated.”} Parents also highlighted learning benefits such as navigating new digital tools, adding structure to story writing, and gaining new English vocabulary. They agreed the system could enhance writing skills, vocabulary, and pronunciation through narration. Overall, parents valued the system’s educational potential and meaningful engagement, while pointing to design and presentation as areas for improvement. These findings suggest that the system may support and encourage learning motivation, though direct educational outcomes were not measured in the present study.

\section{Conclusion \& Future Work}
We presented a multimodal multi-agent system for generating high-quality stories for school children. Our evaluation results suggest strong potential across all media types and highlight steps toward a more robust pipeline. The tool provides an immersive educational experience that is both entertaining and pedagogically valuable. Future work includes integrating image inputs (e.g., children’s drawings) as prompts, adding distinct voices for character dialogues to enhance immersion, and incorporating AudioGen \citep{copet2023simple} as a Foley Artist to generate context-appropriate sound effects behind the animations. In addition, improving the quality of the TTS and TTV modules, as highlighted in the user study, remains a key goal for enhancing the overall storytelling experience. 

Beyond the application itself, our study provides a comparative qualitative evaluation of multiple generative models across three media modalities (text, visual, and audio), highlighting their relative strengths, weaknesses, and capability differences. This evaluation framework can serve as a reference benchmark for assessing future multimodal storytelling systems and related generative models. Moreover, the human-in-the-loop storytelling process explored in this work provides a foundation for future research on AI-assisted data collection and human-AI co-creative educational systems.


\section{Limitations}
The system has several limitations that need to be addressed to enhance its performance and effectiveness. One major limitation is the need for a more comprehensive dataset to robustly test the child-appropriateness of the generated content. Additionally, the generated animations, while not inappropriate, often suffer from visual distortions and lack the smoothness and coherence required for engaging and child-friendly storytelling. These visual inconsistencies can detract from the immersive experience and may not fully align with the intended narrative quality. In addition, the creativity of the generated models is also limited by the training dataset. This may be improved by using a wider genre of stories rather than a large dataset with many similar stories. The fairytales and folktales from Project Gutenberg that were used to train the Writer LLM share many similar themes and plots. This necessitates the creation of a newer, diverse dataset of stories. Another significant limitation is the high generation time for animations, with each six-second video requiring approximately three minutes to render. In our study, the practical impact of this limitation was mitigated for younger participants (ages 6-9). The combination of multiple media elements (text, audio, and visual) increased their content-consumption time, giving animations buffer time to render. However, this is unlikely to generalize to older participants with faster reading speeds. Responsiveness could be significantly improved by reducing video duration or replacing it entirely with static images, replicating a picture-book experience. Evaluating these alternatives constitutes an important direction for future work.

\section{Ethical Impact}
The ethical impact of this system centers on ensuring child safety and fostering positive developmental outcomes. By integrating robust content moderation and focusing on child-appropriate narratives, the system aims to provide a safe storytelling environment. However, ethical considerations include the potential for cultural biases in generated content, requiring careful dataset curation and evaluation to ensure inclusivity and fairness. Additionally, the use of generative AI must prioritize transparency, explicitly informing users about AI-generated content to maintain trust. Addressing these ethical concerns is critical to creating a responsible and impactful educational tool for children.


\bibliography{custom}

\appendix
\section{Implementation Details}
\subsection{Models}
Llama-3.1 models are available under \textit{llama3.1} license and Gemma-2 models are under \textit{gemma} license. GPT-4 is available under proprietary licence. For TTS models, XTTSv2 is under \texttt{coqui-public-model-license} and StyleTTS 2 is using \textit{MIT} license. The TTV model, CogVideoX is using it's own custom license. Finally, MusicGen model uses \textit{CC-BY-NC-4.0}. All models used in this paper comply with their respective license.

\subsection{Dataset}
Our study adheres to the Project Gutenberg License, which allows the use of its texts for non-commercial purposes with proper attribution, ensuring compliance with their terms and conditions.

\subsection{Model Size and Budget}
We use Llama-3.1 in two sizes: 8 billion and 70 billion parameters, and Gemma-2 in two sizes: 9 billion and 27 billion parameters. A single Nvidia A100 80GB GPU was used to deploy the full system.

\subsection{Human Annotators}
There are six human annotators in this study, all of them are undergraduate Computer Science students. All annotators are native speakers of Urdu from Pakistan with English as the medium of education.

\section{System Screen}
\label{sec:screen}
Figure \ref{fig:system-screen} shows the frontend design for the application.

\section{Score Breakdowns}
\label{sec:score-breakdown}
Table \ref{tab:sb-story} gives the score breakdown for story generation, \ref{tab:sb-tts} gives the score breakdown for TTS and \ref{tab:sb-ttv} gives the score breakdown for TTV.

\section{System Prompts}
\label{sec:prompts}
Table \ref{tab:w-r-prompt} shows the system prompt for the Writer and Reviewer. Table \ref{tab:md-prompt} shows the system prompt for Movie Director and Music Director.

\onecolumn
\begin{longtable}{|p{1.0in}|p{4.6in}|}
\caption{\small{Score breakdown for story evaluation.}} \label{tab:sb-story} \\
\hline
\textbf{Criterion} & \textbf{Breakdown} \\
\hline
\endfirsthead
\hline
\textbf{Criterion} & \textbf{Breakdown} \\
\hline
\endhead
\hline
\endfoot
\hline
\endlastfoot
Grammar & \textbf{0:} The story has more or equal to 5 grammatical errors.\newline
The story contains frequent grammatical mistakes that make it difficult to read. For example, "The boy were running fast" and "She don't likes apples." These errors appear in multiple places, affecting the clarity and quality of the writing.\newline
\textbf{1:} The story has less than 5 but more than or equal to 1 grammatical error. The story has a few grammatical errors. For instance, in one sentence, "Their going to the store" instead of "They're going to the store." These errors are noticeable but don’t significantly disrupt the overall flow of the story.\newline
\textbf{2:} The story has no grammatical errors. The story is grammatically correct throughout. Sentences are properly constructed, with no issues related to subject-verb agreement, punctuation, or spelling, allowing for smooth and error-free reading.
\\
\hline
Linguistic Consistency & 
\textbf{0:} The story frequently uses words of varying difficulty and nature. The story starts with simple sentences like "The cat sat on the mat," but later switches to complex, academic phrases such as "The feline exhibited a propensity for reposing on the textile artifact." This frequent change in language style makes it hard to follow and inconsistent in its tone.\newline
\textbf{1:} The story uses words of varying difficulty and nature at some points in the story. The story generally uses straightforward language, but at certain points, it introduces more advanced or specialized terms, like "gregarious" or "nascent," which disrupt the overall flow and accessibility of the text.\newline
\textbf{2:} The story uses consistent language. Throughout the story, the language remains at a similar level of complexity. Whether describing actions, emotions, or settings, the words are accessible and appropriate for the intended audience, making the story easy to follow and cohesive.
 \\
\hline
Appropriate language & 
\textbf{0:} The story is not appropriate for children. It contains several instances of derogatory or offensive terms, such as racial slurs or profanity, making the story unsuitable for most audiences.\newline
\textbf{1:} The story is generally appropriate for children but contains 1-2 instances of language that may be considered offensive or inappropriate. For example, there may be mild profanity or an offensive stereotype in dialogue, though it does not dominate the story.\newline
\textbf{2:} The story uses entirely appropriate language for children, with no offensive or derogatory terms, making it fully suitable for all audiences.
 \\
\hline
Structural Consistency & 
\textbf{0:} The story has frequent disjoints between sentences or paragraphs. The story jumps from one idea to another without smooth transitions. For instance, one paragraph describes a character walking in the park, and the next abruptly shifts to a completely unrelated topic like cooking dinner, with no connection between the scenes. This makes the narrative feel fragmented and difficult to follow.\newline
\textbf{1:} The story has a few disjoints between sentences or paragraphs. Most of the story flows well, but there are a few instances where the transition between ideas or paragraphs feels abrupt. For example, a conversation between characters might end suddenly, followed by an unconnected action scene. These moments disrupt the overall coherence of the story, but they are infrequent.\newline
\textbf{2:} The story has no disjoint between sentences or paragraphs. The story is structurally consistent, with each sentence and paragraph logically connected to the next. Ideas flow naturally, and transitions between scenes or topics are smooth, making the story easy to follow and cohesive from start to finish.
 \\
\hline
Creativity & 
\textbf{0:} The story lacks creativity and does not engage the reader. The story feels dull and uninspired, with predictable events and flat characters. The plot lacks any unique or imaginative elements, and there’s little emotional engagement, making it hard to stay interested or feel invested in the story. For instance, a generic "hero saves the day" narrative without any twists or depth.\newline
\textbf{1:} The story has some creative elements and does engage the reader at some points. The story has some interesting moments and creative ideas, but it doesn't fully capture your attention. Certain plot points or characters may be intriguing, but other sections feel predictable or lacking in excitement. While it holds your attention at times, it’s not consistently engaging.\newline
2: The story is highly creative and engages the reader throughout. The story is captivating from start to finish, with imaginative ideas, strong character development, and unexpected plot twists. The narrative is fresh and unique, making the reader want to keep turning the pages. For example, a well-crafted fantasy world with complex characters and an unpredictable plot that keeps you fully engaged.
 \\
\hline
Adherence to instructions & 
\textbf{0}: Does not adhere to the prompt at all. The story largely ignores the user's prompt or strays far from the requested parameters. For example, if the prompt asks for a mystery story set in a small town, but the story is about a romance in a futuristic city, it doesn't follow the given instructions.\newline
\textbf{1:} Partially adheres to the prompt. The story follows some elements of the prompt but deviates from others. For example, if the prompt requests a suspenseful crime story, but while the beginning sets up a crime, the rest of the story shifts into a lighthearted adventure, it only partially meets the instructions.\newline
\textbf{2:} Fully adheres to the prompt. The story strictly follows the prompt parameters and delivers exactly what was requested. For instance, if the prompt asks for a fantasy adventure with a heroic quest, the story maintains that genre, plot structure, and tone throughout, closely aligning with the user's instructions.
 \\
\hline
Naturalness & 
\textbf{0:} Does not feel human-written. The story feels robotic or mechanical, with awkward phrasing and unnatural sentence structure. For example, "The boy happy was with the dog. They together play every day." It lacks the flow and nuances typical of human writing, making it obvious that it was generated by a machine.\newline
\textbf{1:} Has some human-like elements but does have other unnatural elements. The story mostly reads as if it was written by a human but includes occasional odd wording or phrasing that disrupts the flow. For example, "She walked with a great hastiness to the room, looking all around her peculiarly." These minor issues make it clear that it wasn’t entirely human-written, but it's still mostly coherent.\newline
\textbf{2:} Feels completely human written. The story flows naturally, with smooth transitions, well-constructed sentences, and appropriate use of language. It feels as if it was written by a human, with no awkward phrasing or mechanical errors. For example, "She hurried down the hallway, her eyes darting from side to side as if searching for something just out of reach."
 \\
\hline
\end{longtable}

\newpage
\begin{longtable}{|p{1.0in}|p{4.6in}|}
\caption{\small{Score breakdown for TTS evaluation.}} \label{tab:sb-tts} \\
\hline
\textbf{Criterion} & \textbf{Breakdown} \\
\hline
\endfirsthead
\hline
\textbf{Criterion} & \textbf{Breakdown} \\
\hline
\endhead
\hline
\endfoot
\hline
\endlastfoot
Clarity and Pauses & 
\textbf{0:} Muffled and unclear speech with high listening effort and poor comprehension.\newline
\textbf{1:} Similar words are unclear with medium listening effort to distinguish and average comprehension. \newline
\textbf{2:} Clear and distinct enunciation of all the words with minimal listening effort and high comprehension.
\\
\hline
Information and Emotion Preservation & 
\textbf{0:} Does not attempt to replicate emotion and tone.\newline
\textbf{1:} Attempts to replicate emotion and tone but does not accurately do so. \newline
\textbf{2:} Emotions and tones are correctly replicated and convey the meaning of the text.
\\
\hline
Intonation and Naturalness & 
\textbf{0:} Changes in pitch are lacking and the audio tone is flat. \newline
\textbf{1:} Changes in pitch are not smooth and feel unnatural. \newline
\textbf{2:} Changes in pitch are smooth and natural.
\\
\hline
Fluency and Pronunciation & 
\textbf{0:} Pronunciation is inaccurate and speech is not comprehensible. \newline
\textbf{1:} Pronunciation is inaccurate but words are still correctly comprehensible. \newline
\textbf{2:} Pronunciation is correct and speech is fluent.
\\
\hline
Voice Cloning & 
\textbf{0:} The cloned voice is not recognizable, with noticeable differences in tone, pitch, or speech pattern, and can not be identified as the intended voice.\newline
\textbf{1:} The cloned voice is recognizable but with noticeable differences in tone, pitch, or speech pattern, requiring some effort to identify it as the intended voice.\newline
\textbf{2:} The cloned voice is almost identical to the original, with a consistent tone, pitch, and speech pattern, making it indistinguishable from the intended voice.
\\
\hline
\end{longtable}

\newpage
\begin{longtable}{|p{1.0in}|p{4.6in}|}
\caption{\small{Score breakdown for TTV evaluation.}} \label{tab:sb-ttv} \\
\hline
\textbf{Criterion} & \textbf{Breakdown} \\
\hline
\endfirsthead
\hline
\textbf{Criterion} & \textbf{Breakdown} \\
\hline
\endhead
\hline
\endfoot
\hline
\endlastfoot
Naturalness Assessment & 
\textbf{0:} Noticeable unnatural behaviors, such as jerky movements, odd deformations, or elements that don't align with reality.\newline
\textbf{1:} Some minor oddities but they don't significantly disrupt realism (e.g., slight unnatural movement or minor distortions). \newline
\textbf{2:} No visible unnatural movements or distortions. Everything appears completely realistic.
\\
\hline
Temporal Quality & 
\textbf{0:} Transitions between frames are abrupt and incoherent, with noticeable stutters and jarring jumps. \newline
\textbf{1:} Minor disruptions in the smoothness, but the overall video is still fairly coherent. \newline
\textbf{2:} Transitions between frames are smooth and coherent, with no jarring jumps or stutters.
\\
\hline
Intonation and Naturalness & 
\textbf{0:} Changes in pitch are lacking and the audio tone is flat. \newline
\textbf{1:} Changes in pitch are not smooth and feel unnatural. \newline
\textbf{2:} Changes in pitch are smooth and natural.
\\
\hline
Fine-Grained Alignment & 
\textbf{0:} Major misalignments, such as wrong colors, incorrect motion directions, or speeds that don't match the description.\newline
\textbf{1:} Most attributes align with the prompt, but there are minor discrepancies (e.g., slightly off colors or speed).\newline
\textbf{2:} All attributes align with the prompt, with no discrepancies.
\\
\hline
Overall Alignment & 
\textbf{0:} The video does not align well with the prompt, with significant differences between the described content and what’s shown.\newline
\textbf{1:} Most of the video aligns with the text prompt, but some parts deviate or are missing.\newline
\textbf{2:} The video fully matches the content of the text prompt in terms of overall structure and actions.
\\
\hline
Child-Friendliness & 
\textbf{0:} The video content is not suitable for children (e.g., contains inappropriate visuals or themes).
.\newline
\textbf{1:} The video content is generally appropriate for children, but some minor elements may require supervision or caution.\newline
\textbf{2:} The video content is highly appropriate for children, with no inappropriate visuals and engaging content for the intended age group.
\\
\hline
\end{longtable}

\onecolumn
\begin{table}[h!]
\centering
\caption{\small{System prompt for Writer and Reviewer.}}
\label{tab:w-r-prompt}
\renewcommand{\arraystretch}{1.5}
\setlength{\tabcolsep}{10pt}
\begin{tabular}{|p{1.0in}|p{4.6in}|}
\hline
\textbf{Model} & \textbf{System Prompt} \\
\hline
Writer & Write a folktale or fairytale for children aged 7 to 12 (3rd to 6th graders), based on the story descriptions provided by the user for Propp's narrative functions for five of the Freytag's pyramid layer. The story should fit within 5 paragraphs. Output only a coherent story, without including anything else, such as a title.
\\
\hline
Reviewer & You are a content moderator. Your task is to review the given story. The story should be appropriate for children of age group 7 to 12 (3rd to 6th graders).\newline
Always answer in the following format:\newline
\#\#\# Reasoning:\newline
...add reasoning here...\newline\newline
\#\#\# Is Appropriate: True/False"\\
\hline
Reviewer Update & Your task is to make the given story child-friendly, age group 7 to 12 (3rd to 6th graders). Make upades to the story based on the given feedback. Output only a coherent updated story, without including anything else, such as a title.\\
\hline
\end{tabular}
\end{table}

\onecolumn
\begin{table}[ht]
\centering
\caption{\small{System prompt for Movie Director and Music Director.}}
\label{tab:md-prompt}
\renewcommand{\arraystretch}{1.5}
\setlength{\tabcolsep}{10pt}
\begin{tabular}{|p{1.0in}|p{4.6in}|}
\hline
\textbf{Model} & \textbf{System Prompt} \\
\hline
Movie Director & You’ll be given a paragraph from a story. Your task is to pick ONE part from the paragraph and write a prompt for a text-to-video model. The prompt must contain only ONE motion or action. The prompt must include all relevant objects, describe the environment scene, and describe the characters in the scene. For each paragraph given by the user keep the character description and the environment description consistent. Include motion in the prompt e.g. walking/running, talking, gesturing, interacting with objects, etc. Always start with "In a cartoon/anime/animated world,". \newline \newline
Example Outputs:\newline
"In a cartoon/anime/animated world, a suited astronaut, with the red dust of Mars clinging to their boots, reaches out to shake hands with an alien being, their skin a shimmering blue, under the pink-tinged sky of the fourth planet. In the background, a sleek silver rocket, a beacon of human ingenuity, stands tall, its engines powered down, as the two representatives of different worlds exchange a historic greeting amidst the desolate beauty of the Martian landscape."\newline\newline
"In a cartoon/anime/animated world, a garden comes to life as a kaleidoscope of butterflies flutters amidst the blossoms, their delicate wings casting shadows on the petals below. In the background, a grand fountain cascades water with a gentle splendor, its rhythmic sound providing a soothing backdrop. Beneath the cool shade of a mature tree, a solitary wooden chair invites solitude and reflection, its smooth surface worn by the touch of countless visitors seeking a moment of tranquility in nature's embrace."
\\
\hline
Music Director & You'll be given a paragraph from a story. Your task is generate a music composition for the emotions in the scene of the story. Make sure to output short one-sentence composition just like the ones given in example outputs. The composition should be simple (like in examples) and ONLY describe the music.\newline \newline
Example Outputs:\newline
"Whimsical orchestral piece with playful flutes, light strings, and occasional harp glissandos."\newline \newline
"Melancholic piano melody with soft strings, gradually building to a heartfelt crescendo."\newline \newline
"Epic orchestral track with powerful brass, thunderous drums, and intense string staccatos."\newline \newline
"Warm, gentle strings with plucked notes, accompanied by a soft flute melody"\\
\hline
\end{tabular}
\end{table}

\end{document}